\documentclass[10pt,twocolumn,letterpaper]{article}

\usepackage{cvpr}              

\usepackage{graphicx}
\usepackage{amsmath}
\usepackage{amssymb}
\usepackage{booktabs}
\usepackage{multirow}
\usepackage{booktabs}
\usepackage{tabularx}
\usepackage{makecell}
\usepackage{multirow}
\usepackage{makecell}
\usepackage{marvosym}
\usepackage{capt-of}
\usepackage[accsupp]{axessibility}
\usepackage{multicol}
\usepackage{blindtext}

\newcolumntype{Y}{>{\centering\arraybackslash}X}

\usepackage{stackengine}

\usepackage[pagebackref,breaklinks,colorlinks]{hyperref}

\usepackage[capitalize]{cleveref}
\crefname{section}{Sec.}{Secs.}
\Crefname{section}{Section}{Sections}
\Crefname{table}{Table}{Tables}
\crefname{table}{Tab.}{Tabs.}

\def\cvprPaperID{2410} 
\def\confName{CVPR}
\def\confYear{2022}

\begin{document}

\title{Pop-Out Motion: 3D-Aware Image Deformation \\via Learning the Shape Laplacian}

\author{Jihyun Lee*$^1$
\qquad
Minhyuk Sung*$^{\dag1}$
\qquad
Hyunjin Kim$^1$
\qquad
Tae-Kyun Kim$^{1,2}$\\
$^1$ KAIST \qquad $^2$ Imperial College London}

\twocolumn[{%
\renewcommand\twocolumn[1][]{#1}%
\maketitle
\begin{center}
\centering
\captionsetup{type=figure}
\vspace{-0.6cm}
\includegraphics[width=\textwidth]{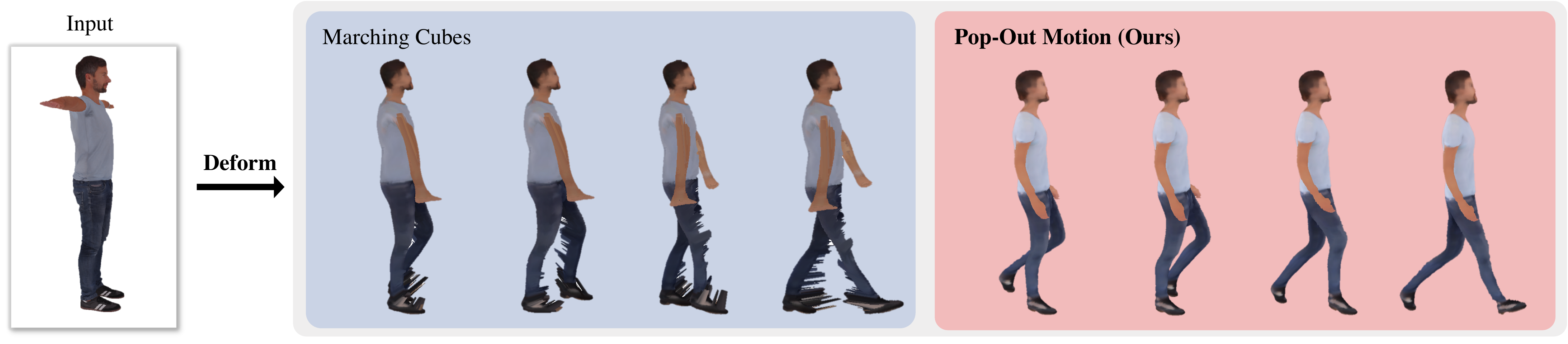} \vspace{-0.6cm}

\caption{\textbf{3D-aware image deformation of a 2D human image.} Compared to the images generated by directly deforming a mesh reconstructed using PIFu \cite{saito2019pifu} and Marching Cubes \cite{lorensen1987marching} (blue), our method (red) can produce more plausible 3D-aware image deformations via inferring an additional intrinsic shape property: the shape Laplacian.}
\label{fig:teaser_image}
\end{center}
\vspace{0.2cm}
}]

{\let\thefootnote\relax\footnote{{* equal contributions,\, \dag\, corresponding author}}}

\setcounter{footnote}{0}

\vspace{-0.5cm}
\begin{abstract}
\vspace{-0.5\baselineskip}
We propose a framework that can deform an object in a 2D image as it exists in 3D space. Most existing methods for 3D-aware image manipulation are limited to (1) only changing the global scene information or depth, or (2) manipulating an object of specific categories. In this paper, we present a 3D-aware image deformation method with minimal restrictions on shape category and deformation type. While our framework leverages 2D-to-3D reconstruction, we argue that reconstruction is not sufficient for realistic deformations due to the vulnerability to topological errors. Thus, we propose to take a supervised learning-based approach to predict the shape Laplacian of the underlying volume of a 3D reconstruction represented as a point cloud. Given the deformation energy calculated using the \emph{predicted} shape Laplacian and user-defined deformation handles (e.g., keypoints), we obtain bounded biharmonic weights to model plausible handle-based image deformation. In the experiments, we present our results of deforming 2D character and clothed human images. We also quantitatively show that our approach can produce more accurate deformation weights compared to alternative methods (i.e., mesh reconstruction and point cloud Laplacian methods).

\end{abstract}

\vspace{-0.4cm}

\section{Introduction}
\label{sec:intro}

The capability of photo editing, which had been confined in 2D space, has recently been popped out into 3D space. For example, predicting depth from a 2D image enables the composition of objects in an image ~\cite{Schnyder:2012,Ziegler:2015}. Object segmentation allows the projection of an image to a new view~\cite{Lee:2014,Xian:2019,Niklaus:2019}. Lighting and ground plane estimation make it possible to relight objects and generate a new shadow in an image ~\cite{Geoffroy:2017,Geoffroy:2018,Zhang:2019,Gardner:2019}. Such techniques for 3D-aware image editing have allowed the user to manipulate an image in a more intuitive manner --- as if the object exists in 3D space --- and opened new opportunities in downstream applications.

As a missing piece of the existing 3D-aware image manipulation methods, we focus on \emph{3D-aware image deformation}. Unlike the aforementioned techniques, 3D-aware deformation does not just alter the scene information (e.g., camera parameters, lighting conditions) or modify the 2.5D information. Instead, it allows the user to directly manipulate the 3D geometry and appearance of an object. More relevant topics to 3D-aware deformation would be (1) human pose transfer~\cite{Ma:2017,Li:2019,Chan:2019}, which works only for human bodies, (2) novel view synthesis~\cite{Wu:2020,Pan:2021,hoiem2005automatic}, which is limited to altering a viewpoint of an image, and (3) 3D model-based manipulation~\cite{Kholgade:2014}, which requires the exact 3D model of the object in an image. To address these limitations, we aim to enable 3D-aware image deformation with minimal restrictions on shape category and deformation type.

For 3D-aware deformation, it is \emph{necessary} to reconstruct the object in a 2D image to 3D space; however, it is \emph{not sufficient} in general. Deformation requires either a surface or volume information~\cite{jacobson2011bounded,wang2015linear}. However, most of existing methods of image-based 3D reconstruction do not directly output a surface or volume ~\cite{Fan:2017,Wang:2019} or produce a surface without proper consideration about intrinsic shape properties~\cite{saito2019pifu,he2020geo,Saito:2020,Lei:2020,Xu:2019,wang2018pixel2mesh,Wen:2019,Kanazawa:2018} -- which can largely affect the deformation result. See an example in the red branch of Figure~\ref{fig:teaser_image}. The inaccurate topological prediction connecting legs causes undesired visual artifacts in the deformation.
Indeed, aiming for topological correctness in 3D reconstruction is a difficult task due to its nature defined with both continuous and discrete quantities.

In this paper, given a 3D point cloud of the object in an input image (whose 2D-to-3D reconstruction is performed by PIFu \cite{saito2019pifu}), we propose to enable 3D-aware image deformation through learning the additional intrinsic geometric property: the shape Laplacian. The shape Laplacian is the \emph{essential} information encoding the geometry intrinsics. In particular, bounded biharmonic weights~\cite{jacobson2011bounded} -- which has been widely used as a standard technique for deformation in computer graphics -- compute the linear blending weights associated with deformation handles as the minimizers of the deformation energy defined using the shape Laplacian. In our framework, we utilize the \emph{estimated} shape Laplacian of a 3D reconstruction to obtain bounded biharmonic weights to plausibly model handle-based image deformation.

To this end, we introduce a neural network that can predict the shape Laplacian of the underlying volume of a 3D point cloud reconstructed from a 2D image --- without directly converting the point cloud to a volume. Considering that the deformation energy can be discretized with the standard linear FEM Laplacian $LM^{-1}L$ (where $L$ is a symmetric cotangent Laplacian matrix and $M$ is a diagonal lumped mass matrix), we design our network to learn the matrices $L$ and $M^{-1}$ from the supervision obtained from a ground truth 3D mesh. The elements in the inverse mass matrix $M^{-1}$ are predicted for each individual point, while the elements of the cotangent Laplacian matrix $L$ are predicted by taking \emph{pairs} of the input points. We use a symmetric feature aggregation function for such pairs and also a weight module to enforce the output matrix $L$ to be symmetric and sparse. In test time, we recover the deformation energy from the predicted $L$ and $M^{-1}$ to compute bounded biharmonic weights with user-specified deformation handles. Since our method learns the shape Laplacian instead of the handle-dependent deformation weights, it can generalize well to arbitrary handle configurations.

In the experiments, we show our results of 3D-aware deformation on 2D character and clothed human images. We also showcase an user-interactive image editing scenario, where the user produces intuitive 3D deformations based on the specified control points. For quantitative evaluation, we test our method on a large-scale 3D point cloud dataset (i.e., DFAUST \cite{dfaust:CVPR:2017}), in which our method is shown to produce more accurate deformation weights compared to the alternative methods on mesh reconstruction and point cloud Laplacian. 

Our main contributions can be summarized as follows:

\vspace{-0.5\baselineskip}

\begin{itemize}

\item We propose a method for 3D-aware deformation of 2D images, which can be applied with minimal restrictions on shape category and deformation type.

\vspace{-0.3\baselineskip}

\item We introduce a novel network architecture that can learn the shape Laplacian with several desired properties (i.e., positive semi-definiteness, symmetry and sparsity) from a 3D reconstruction. To the best of our knowledge, this is the first study to demonstrate that a learning-based approach can be effective in predicting the shape Laplacian of the underlying volume of a point cloud.

\vspace{-0.3\baselineskip}

\item We empirically demonstrate that our \emph{learning}-based approach leads to more plausible deformations compared to the alternative cases of \emph{calculating} the approximation of the shape Laplacian using mesh reconstruction or point cloud Laplacian methods.

\end{itemize}
\section{Related Work}
\label{sec:related_work}

\paragraph{Handle-Based Deformation} In geometry processing, methods for handle-based deformation have been investigated for decades. A common workflow for handle-based deformation consists of two phases: bind time and pose time. In bind time, the source shape is bound to a user-defined set of control handles. In pose time, the handles are manipulated to produce shape deformations in a way that the transformations at each handle are smoothly propagated to the rest of the shape. Most of existing methods compute handle-based deformation weights based on solving an optimization problem to minimize a shape fairness functional (e.g., discretized Laplacian energy \cite{jacobson2011bounded}, linearly precise smoothness energy \cite{wang2015linear}). In this work, we utilize bounded biharmonic weights \cite{jacobson2011bounded} to model handle-based deformation for 2D images.

\vspace{-0.7\baselineskip}

\paragraph{Learning Shape Deformation} Most of recent methods in 3D vision and graphics utilize neural networks to learn shape deformations. Typically, the main objective of such existing work is to fit the source to the target shape. These target-driven deformation methods have been shown to be effective in various tasks, such as 3D reconstruction \cite{wang2018pixel2mesh,jack2018learning,wang20193dn}, shape auto-encoding \cite{tretschk2020demea}, deformation transfer \cite{sung2020deformsyncnet}, and data augmentation \cite{atzmon2021augmenting,muralikrishnan2021glass}. Unlike these methods, the main goal of our work is to compute deformation weights associated with \emph{user-defined control handles} -- which can serve as an intuitive deformation interface -- for a 3D reconstruction whose ground truth topology is unknown. We also note that our framework does not require any source and target shape pairs or semantic labels for training.

\vspace{-0.7\baselineskip}

\paragraph{Mesh Reconstruction Methods}
One of the alternative approaches to compute handle-based deformation weights for a 3D point cloud is to compute the shape Laplacian using the mesh topology estimated from a surface reconstruction method, an optional manifold conversion algorithm \cite{huang2018robust,huang2020manifoldplus}), and a tetrahedral meshing method \cite{hu2020fast}. Explicit surface reconstruction methods \cite{bernardini1999ball,boissonnat1993three,liu2020meshing} directly estimate the connectivity information for input points. Implicit surface reconstruction methods \cite{kazhdan2006poisson,kazhdan2013screened,park2019deepsdf} predict a field function that can be used for iso-surface meshing, which is typically performed via Marching Cubes \cite{lorensen1987marching}. However, most of these existing methods are vulnerable to topological errors when ambiguous structures (e.g., spatially adjacent surfaces, high-curvature surface) exist. In this paper, we propose to bypass such explicit mesh conversion procedure and to directly learn the shape Laplacian of a point cloud from the ground truth topology supervision. In Section \ref{sec:results}, we empirically demonstrate that our method can produce more accurate deformation weights than the aforementioned scenario.

\vspace{-0.3\baselineskip}

\paragraph{Point Cloud Laplacians} Another alternative approach to calculate handle-based deformation weights for a reconstructed 3D point cloud is to directly construct a point cloud Laplacian. While there are existing methods that approximate the shape Laplacian of a point cloud \cite{belkin2009constructing,sharp2020laplacian}, the main challenge is to infer the correct topology of the underlying structure of a point cloud -- which is infeasible using the given point cloud only. The construction of a point cloud Laplacian is commonly based on the triangulation on the tangent planes estimated from the $k$ nearest neighbors of each point in the \emph{Euclidean} space. Therefore, the resulting shape Laplacian is erroneous with respect to the two neighboring points with a long underlying \emph{geodesic} distance. To address this issue, we propose to leverage the prediction power of a neural network, which is trained using the \emph{ground truth} shape Laplacian as a supervisory signal. To the best of our knowledge, this is the first study to take a learning-based approach to predict the shape Laplacian of the underlying volume from a point cloud.

\section{Learning for Handle-Based Deformation}
\label{sec:method}

We propose a learning-based method for 3D-aware image deformation.
First, we use a 3D reconstruction method (i.e., PIFu \cite{saito2019pifu}) to construct a point cloud that represents the 3D shape of the object in an input image. Next, we use a carefully-designed neural network to predict the shape Laplacian of the underlying volume of the reconstructed 3D point cloud. The estimated shape Laplacian is then used to compute the deformation blending weights (i.e., bounded biharmonic weights \cite{jacobson2011bounded}) corresponding to user-defined handles to model 3D-aware image deformation. We argue that our approach to \emph{learning} the shape Laplacian allows more robust and accurate deformations than directly \emph{calculating} the approximation of the shape Laplacian using a mesh reconstruction or point cloud Laplacian method (please refer to Section~\ref{sec:results} for experimental results). In addition, since we predict the shape Laplacian -- not directly the deformation blending weights that are dependent on deformation handles, our method can model deformation robust to an arbitrary deformation handle configuration.

In what follows, we first briefly review bounded biharmonic weights \cite{jacobson2011bounded}, which is one of the existing methods to compute handle-based deformation weights for \emph{a mesh}, whose ground truth topology is available. We then introduce our method to obtain robust bounded biharmonic weights for a 3D reconstruction.

\subsection{Background: Bounded Biharmonic Weights}
\label{sec:BBW}


Bounded biharmonic weights~\cite{jacobson2011bounded} are linear blending weights that propagate affine transformations defined at arbitrary control handles to the other points in the shape.
Given a tetrahedral mesh\footnote{While it is also possible to compute the shape Laplacian from a \emph{surface} mesh, we focus on the \emph{volume} Laplacian in this work.} $\mathcal{M} = \{\mathcal{V}, \mathcal{F}\}$ representing the volume of a shape with the sets of vertices and faces $\mathcal{V}$ and $\mathcal{F}$, respectively, a control handle $\mathcal{H}$ is defined as a point ($\mathcal{H} \in \mathcal{V}$) or a region ($\mathcal{H} \subset \mathcal{V}$) on the mesh\footnote{The handle can also be defined \emph{out of} the mesh as a skeleton bone or a vertex of a cage. Here, we only describe the case when the handle is defined as a point or a region on the mesh.}.
Specifically, given (1) the source shape $\mathcal{M}$ with $n$ vertices ($\mathcal{V}=\{ \mathbf{v}_i \}_{i=1 \cdots n}$) (2) $m$ number of control handles $\{ \mathcal{H}_k \}_{k=1 \cdots m}$, and (3) the affine transformations $\{ \mathbf{T}_k \}_{k=1 \cdots m}$ defined on each control handle, the new position of $i$-th vertex $\mathbf{v}_i \in \mathcal{V}$ can be calculated using the following linear formulation:

\vspace{-0.4cm}
\begin{align}
\mathbf{v}_i' = \sum_{k=1}^{m} w_{k,i} \mathbf{T}_k \mathbf{v}_i,
\end{align}

\noindent where $w_{k,i}$ is a deformation weight associated with the $k$-th control handle $\mathcal{H}_k$ and the $i$-th vertex $\mathbf{v}_i$.

The deformation weights $\mathbf{w}_k = \{w_{k,1}, \cdots, w_{k,n}\}^T$ for each handle are computed as a minimizer with respect to the deformation energy subject to several constraints for the desired properties of deformation (e.g., partition of unity, non-negativity). The minimization problem can be written as follows:

\vspace{-0.45cm}
\begin{align}
\underset{\{ \mathbf{w}_k \}_{k=1 \cdots m}}{\mathop{\mathrm{argmin}}}& \sum_{k=1}^{m} \frac{1}{2}\; \mathbf{w}_k^T\, A\, \mathbf{w}_k
\end{align}
\label{eq:bbw}
\vspace{-0.5cm}
{\footnotesize
\begin{align}
    \qquad\qquad\text{subject to: }\; & w_{k,i} = 1 \quad \forall i \quad \text{s.t.} \quad \mathbf{v}_i \in \mathcal{H}_k  \nonumber\\
    & w_{k,i} = 0 \quad \forall i \quad \text{s.t.} \quad \mathbf{v}_i\in \mathcal{H}_{l, l \neq k}  \nonumber\\
    & \textstyle \sum_{k=1}^{m} w_{k,i}=1, \enspace i=1,\cdots,n, \nonumber\\
    & 0 \leq w_{k,i} \leq 1, \enspace k=1,\cdots,m, \enspace i=1,\cdots,n, \nonumber
\end{align}}
\vspace{-0.3cm}

\begin{figure*}[!t]
\begin{center}
\vspace{-0.2\baselineskip}
\includegraphics[width=\textwidth]{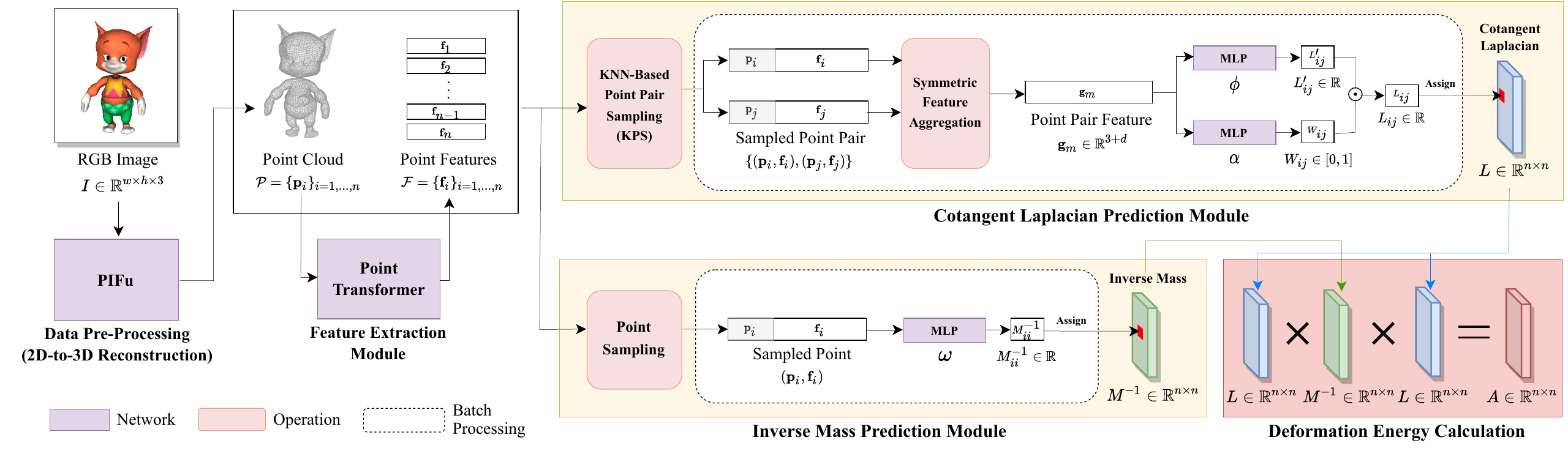} 
\end{center}
\vspace{-1.3\baselineskip}

\caption{\textbf{Architecture overview.} Given a point cloud $\mathcal{P} = \{ \mathbf{p}_i \}_{i = 1 \cdots n}$ (i.e., a 3D reconstruction of an RGB image $\mathcal{I}$), our method learns the shape Laplacian to compute the deformation energy matrix $A \in \mathbb{R}^{n \times n}$ of the underlying volume of $\mathcal{P}$. The proposed framework mainly consists of three modules: (1) Feature Extraction Module, (2) Cotangent Laplacian Prediction Module, and (3) Inverse Mass Prediction Module. Feature Extraction Module first extracts per-point features $\mathcal{F} = \{ \mathbf{f}_i \}_{i = 1 \cdots n}$ of $\mathcal{P}$. Given $\mathcal{P}$ and $\mathcal{F}$, Cotangent Laplacian Prediction Module estimates the cotangent Laplacian matrix $L \in \mathbb{R} ^ {n \times n}$ and Inverse Mass Prediction Module predicts the inverse mass matrix $M^{-1} \in \mathbb{R} ^ {n \times n}$ of $\mathcal{P}$. The final deformation energy $A$ can be obtained as $LM^{-1}L$.}
\label{fig:overall_architecture}
\vspace{-0.75\baselineskip}
\end{figure*}
\noindent where $A=LM^{-1}L$ is the deformation energy matrix of the source mesh $\mathcal{M}$, $L \in \mathbb{R}^{n \times n}$ is the cotangent Laplacian matrix of $\mathcal{M}$, and $M \in \mathbb{R}^{n \times n}$ is a lumped mass matrix whose diagonal elements represent the volume associated with each of the vertices in $\mathcal{M}$. Both $L$ and $M$ are defined on \emph{volume} information of a shape and thus cannot be directly computed from a point cloud. Although it is possible to recover its volume using a mesh reconstruction method (e.g., \cite{bernardini1999ball,boissonnat1993three,kazhdan2006poisson,kazhdan2013screened,liu2020meshing,park2019deepsdf,hu2020fast}), it may introduce topological noises that result in erroneous deformation weights. To address this issue, we argue that \emph{learning} for the deformation energy $A$ enables more accurate and robust deformation.

\subsection{Laplacian Learning Network}
\label{subsec:laplacian_learning_network}

We now introduce our network that can learn the deformation energy matrix $A \in \mathbb{R} ^ {n \times n}$ of the underlying volume of a 3D reconstruction represented as a point cloud $\mathcal{P} = \{ \mathbf{p}_i \}_{i = 1 \cdots n}$. Instead of directly estimating the matrix $A$, we propose to learn the cotangent Laplacian matrix $L \in \mathbb{R}^{n \times n}$ and the inverse mass matrix $M^{-1} \in \mathbb{R}^{n \times n}$ such that the deformation energy can be later recovered as $A = LM^{-1}L$. The first motivation behind this design is to automatically guarantee the positive semi-definiteness of the predicted $A$; it is non-trivial to enforce this property when directly predicting a squared matrix using a neural network. The second motivation is to better learn and enforce the symmetry and sparsity structure of the cotangent Laplacian matrix $L$ to allow a more accurate prediction.


In what follows, we explain the detailed architecture of our Laplacian learning network, which consists of three modules: (1) Feature Extraction Module, (2) Cotangent Laplacian Prediction Module, and (3) Inverse Mass Prediction Module. 

\vspace{0.1cm}
\noindent \textbf{Feature Extraction Module} Given a point cloud $\mathcal{P} = \{ \mathbf{p}_i \}_{i = 1 \cdots n}$, Feature Extraction Module extracts a point cloud feature $\mathcal{F} = \{ \mathbf{f}_i \}_{i = 1 \cdots n}$, where $\mathbf{f}_i  \in \mathbb{R} ^ d$ denotes a per-point feature vector corresponding to $\mathbf{p}_i$. For network architecture, we adopt Point Transformer \cite{zhao2021point} that can extract permutation and cardinality-invariant feature of a point cloud via self-attention operations. 

\vspace{0.1cm}
\noindent \textbf{Cotangent Laplacian Prediction Module} Given the point cloud $\mathcal{P}$ and the point cloud feature $\mathcal{F}$, Cotangent Laplacian Prediction Module learns the cotangent Laplacian matrix $L \in \mathbb{R} ^ {n \times n}$ of the underlying volume of $\mathcal{P}$. Following the definition of the cotangent Laplacian, $L$ is desired to be a symmetric matrix whose element $L_{ij}$ is non-zero only if $\mathbf{p}_i$ and $\mathbf{p}_j$ are topologically connected by an edge. Since $L$ is highly sparse, and the point cloud cardinality $n$ is typically large (e.g., in the order of thousands), it is inefficient to learn the relationship between \emph{all} point pairs. Therefore,
we use a Euclidean prior to select the initial point pair candidates which are probable to have local connectivity -- thus correspond to \emph{non-zero} entries of $L$. More specifically, for each point $\mathbf{p}_i \in \mathcal{P}$, we compute $k$ nearest neighbor points based on Euclidean distance. We then couple each of the neighbor points with the source point $\mathbf{p}_i$ to form point pair candidates. We denote this sampling strategy by \emph{KNN-Based Point Pair Sampling (KPS)}. We note that using KPS not only improves the inference time but also helps the network training by alleviating an imbalanced regression problem, as a significant number of zero-valued regression targets can be filtered out before training. In our experiments, we empirically set $k$ as 32.

We now extract a feature for each point pair candidate as follows:

\vspace{-0.2cm}
{\small
\begin{equation}
    \mathbf{g}_{m} = \left( \gamma_1(\mathbf{p}_i,\, \mathbf{p}_j), \gamma_2(\mathbf{f}_i, \mathbf{f}_j) \right).
\end{equation}}
\vspace{-0.2cm}

\noindent where $\{\mathbf{p}_i,\, \mathbf{p}_j\}$ is the $m$-th point pair candidate, and $\gamma_1(\cdot)$ and $\gamma_2(\cdot)$ are \emph{symmetric} functions used for the pairwise feature aggregation. Since the matrix $L$ is desired to be \emph{symmetric}, the pairwise feature aggregations must be symmetric to produce the identical features for inputs $(\mathbf{p}_i, \mathbf{p}_j)$ and $(\mathbf{p}_j, \mathbf{p}_i)$ -- thus guaranteeing the equality between the later predictions for $L_{ij}$ and $L_{ji}$. We empirically choose absolute difference and element-wise multiplication for $\gamma_1$ and $\gamma_2$, respectively. (Refer to Section~\ref{sec:results} for the ablation study.)

Next, we estimate the entry of the cotangent Laplacian $L$ that corresponds to each point pair candidate as follows:

\vspace{-0.1cm}
\begin{equation}
    L_{ij} = \alpha(\mathbf{g}_{m}) \odot \phi(\mathbf{g}_{m}),
\end{equation}
\vspace{-0.3cm}

\noindent where $i$ and $j$ are the indices of the $m$-th point pair candidate, $\phi(\cdot)$ is a function that outputs a real-valued scalar, and $\alpha(\cdot)$ is a function predicting a \emph{weight} $W_{ij} \in [0, 1]$ indicating whether the element $L_{ij}$ in the cotangent Laplacian is a non-zero value.
The addition of the weight prediction is our \emph{key} to improving the accuracy of the deformation.
As previously mentioned, $L$ has non-zero elements only for the point pairs that are connected through edges; this indicates that $L$ is a very sparse matrix. Capturing its sparsity structure is essential, since it encodes information about the topology of the shape. To better model such sparsity structure, we introduce the function $\alpha(\cdot)$ and predict the weight indicating whether each element in $L$ is zero or not. We empirically find that this additional sparsity structure prediction significantly improves the quality of deformations, as shown in Section~\ref{sec:results}.
We also remark that our architecture of the Cotangent Laplacian Prediction Module --- taking point pairs as input --- has the advantage of learning well even from a scarce training dataset since the number of training examples increases by $k$ with the number of sampled points.

In test time, the entire cotangent Laplacian matrix $L$ can be computed by predicting the off-diagonal elements $\{ L_{ij} \}_{i \neq j}$ in parallel via batch processing and calculating diagonal elements $\{ L_{ii} \}_{i = 1,\cdots,n}$ as the minus sum of the off-diagonal elements in a row -- to follow the definition of the cotangent Laplacian.

\vspace{0.2cm}
\noindent \textbf{Inverse Mass Prediction Module} Given the point cloud $\mathcal{P}$ and the point cloud feature $\mathcal{F}$, Inverse Mass Prediction Module learns the inverse mass matrix $M^{-1} \in \mathbb{R}^{n \times n}$ of the underlying volume of $\mathcal{P}$. For each point in $\mathcal{P}$, we first concatenate its 3D coordinate $\mathbf{p}_i$ and the corresponding feature $\mathbf{f}_i$ to generate a new per-point feature $\mathbf{f'}_i = [\mathbf{p}_i; \mathbf{f}_i] \in \mathbb{R} ^ {3 + d}$. Then, the inverse mass of each point $\mathbf{p}_i$ can be predicted as $\omega(\mathbf{f'}_i),$ where $\omega$ is a function instantiated as an MLP. In test time, the entire inverse mass matrix $M^{-1}$ can be computed by predicting the diagonal elements in a batch and padding the off-diagonal entries as zero.

\vspace{0.2cm}
\noindent \textbf{Loss Functions.} We train our network using the supervision obtained from the  ground truth mesh corresponding to the input point cloud $\mathcal{P}$. The overall loss function for our framework can be written as follows:

\vspace{-0.25cm}
{\scriptsize 
\begin{align}
   \mathcal{L} = \sum_{i, j \in \mathcal{K}(\mathcal{P})} \frac{1}{|\mathcal{K}(\mathcal{P})|}&\| L_{ij} - L^{\text{gt}}_{ij} \|_1\ 
   + \sum_{i, j \in \mathcal{K}(\mathcal{P})} \frac{\lambda_{W}}{|\mathcal{K}(\mathcal{P})|}  \| W_{ij} - W^{\text{gt}}_{ij} \|_1 \nonumber \\[0.1pt] 
   &+ \sum_i^{|\mathcal{P}|} \frac{\lambda_{M^{-1}}}{|\mathcal{P}|} \| M^{-1}_{ii} - M^{-1\text{gt}}_{ii} \|_1, 
\end{align}}
\vspace{-0.25cm}

\noindent where $L^{\text{gt}}$, $W^{\text{gt}}$, and $M^{-1\text{gt}}$ are ground truth matrices for $L$, $W$ and $M^{-1}$, respectively. $\mathcal{K}(\mathcal{P})$ denotes a set of point pair indices sampled by KPS. For the weights of the loss terms $\lambda_W$ and $\lambda_{M^{-1}}$, we empirically choose 100 and 1, respectively.

\begin{table*}[!t]
\centering
{ \scriptsize	
\setlength{\tabcolsep}{0.2em}
\renewcommand{\arraystretch}{1.0}
\caption{\textbf{Quantitative comparison of 3D point cloud deformation results on DFAUST \cite{dfaust:CVPR:2017} dataset.} The point handles for a source shape are selected using farthest point sampling.}
\vspace{-0.15\baselineskip}
\label{table:dfaust_results}
\begin{tabularx}{\textwidth}{>{\centering}m{1.2cm}|>{\centering}m{2.3cm}|YYYYYYYYY}
\toprule
Number of Handles &
Metric &

PSR \cite{kazhdan2013screened} &
APSS \cite{apss} &
BPA \cite{bernardini1999ball} &
DeepSDF \cite{park2019deepsdf} &
DGP \cite{williams2019deep} & 
MIER \cite{liu2020meshing} &
PCDLap \cite{belkin2009constructing} &
NMLap \cite{sharp2020laplacian} & 
Ours \\

\midrule
\multirow{3}{*}{16} & Weight L1 ($\times$100) $\downarrow$ & 3.86 & 3.46 & 4.32 & 2.66 & 4.15 & 3.26 & 3.53 & 3.34 &\textbf{2.10} \\
& Shape CD ($\times$100) $\downarrow$ & 3.84 & 3.04 & 3.83 & 2.61 & 4.09 & 3.16 & 2.97 & 4.04 & \textbf{1.81} \\
& Shape HD ($\times$0.1) $\downarrow$ & 1.81 & 1.31 & 1.73 & 0.48 & 2.85 & 1.13 & \textbf{0.42} & 0.43 & \textbf{0.42} \\
 
\midrule
\multirow{3}{*}{32} & Weight L1 ($\times$100) $\downarrow$  & 3.08 & 1.53 & 1.72 & 1.38 & 3.01 & 1.09 & 1.54 & 1.49 &\textbf{1.06} \\
& Shape CD ($\times$100) $\downarrow$ & 5.12 & 2.39 & 2.29 & 2.15 & 4.55 & 1.26 & 2.06 & 2.18 & \textbf{1.45} \\
& Shape HD ($\times$0.1) $\downarrow$  & 3.39 & 0.79 & 0.83 & 0.57 & 4.25 & 2.10 & \textbf{0.53} & \textbf{0.53} & \textbf{0.53}\\                                                   

\bottomrule
\end{tabularx}
}
\vspace{-0.6\baselineskip}
\end{table*}

\noindent \textbf{Deformation Weight Computation.} 
In test time, once we predict the cotangent Laplacian matrix $L$ and the inverse mass diagonal matrix $M^{-1}$ to recover the deformation energy matrix $A = LM^{-1}L$, the linear deformation blending weights in Equation~\ref{eq:bbw} for the set of deformation handles are calculated by solving the quadratic programming problem. Note that, while the deformation blending weights are \emph{dependent} on the given set of deformation handles, the quantities we predict ($L$ and $M^{-1}$) are \emph{not dependent} on the deformation handles. Hence, our framework can allow the user to choose an arbitrary set of deformation handles and move them freely without any restriction.

\subsection{Implementation Details}
\label{subsec:implementation_details}

\noindent \textbf{Network Architectures.} For Feature Extraction Module, we adopt Point Transformer \cite{zhao2021point} architecture to output a 64-dimensional feature for each point $\mathbf{p}_i$. For $\alpha$ and $\phi$ in Cotangent Laplcian Prediction Module and $\omega$ in Inverse Mass Prediction Module, we use multilayer perceptron (MLP) architecture composed of three fully-connected layers, each of them followed by batch normalization, dropout and LeakyReLU -- except for the last activation. For $\alpha$ and $\omega$, we adopt sigmoid as the final activation, while we use no activation for the last layer of $\phi$. The output feature dimensions for each of the layers are 128, 256, and 1, respectively. We purposely design a lightweight architecture for $\alpha$, $\phi$, and $\omega$, as they are desired to operate in parallel to batch-compute the entire cotangent Laplacian matrix $L$ and the inverse mass matrix $M^{-1}$. Due to the space limit, please refer to supplementary material for more details of our network training (e.g., learning rate, batch size).

\noindent \textbf{Deformation Weight Computation.} We use the implementation of libigl \cite{libigl} to compute bounded biharmonic weights. This implementation solves the quadratic programming problem in Equation \ref{eq:bbw} by executing Mosek \cite{mosek}, which is a software package for solving large-scale optimization problems. We also utilize libigl library for computing ground truth cotangent stiffness matrix $L$ and mass matrix $M$ to obtain supervision for our network training.

\section{Experiments}
\label{sec:results}
In this section, we experimentally validate the effectiveness of our method. In Section~\ref{subsec:3d_point_cloud_deformation}, we first evaluate our method on a large-scale 3D point cloud dataset for \emph{quantitative} comparisons with the alternative methods. In Section~\ref{subsec:3d_aware_image_deformation}, we present our results of 3D-aware image deformation. Lastly, we report our ablation study in Section~\ref{subsec:ablation_studies}.

\subsection{3D Point Cloud Deformation}
\label{subsec:3d_point_cloud_deformation}

To \emph{quantitatively} evaluate our deformed shape quality, we first test our method on a large-scale 3D point cloud dataset (i.e., DFAUST~\cite{dfaust:CVPR:2017}). We also report our deformation results on 3D \emph{partial} point clouds (i.e., SHREC'16 \cite{cosmo2016shrec}).
\vspace{-0.3cm}

\subsubsection{Experiments on 3D Point Clouds}
\label{subsec:dfaust_experiments}

In this subsection, we perform quantitative evaluation on DFAUST~\cite{dfaust:CVPR:2017} dataset, which contains 40,000 3D human body meshes that represent 129 motion sequences of 10 real humans. While no image is used in this experiment, we consider the point clouds generated from the meshes are 3D reconstructions for the sake of \emph{quantitative} comparisons. To prepare the data, we first subsample the shapes in each sequence by a factor of 16 due to the high frame rate of the motion sequences. We then use the shapes of eight humans for training and the shapes of the remaining two humans for testing. For evaluation metrics, we use (1) L1 distance (Weight L1) between the ground truth and the predicted deformation weights, (2) Chamfer distance (Shape CD) and (3) Hausdorff distance (Shape HD) between the ground truth and the predicted deformed point clouds. For more detailed information about the metric computation, please refer to our supplementary section.

\begin{figure}[!b]
\vspace{-0.6\baselineskip}
\centering
\setlength{\tabcolsep}{0em}
\def\arraystretch{0.0}
\newcolumntype{Y}{>{\centering\arraybackslash}m{0.09\columnwidth}}
{\tiny
\begin{tabular}{Y|Y|YYYYYYYY|Y}
Input & \makecell{GT\\Deform} 
& \makecell{PSR \\\cite{kazhdan2013screened}} &
\makecell{APSS  \\\cite{apss}} &
\makecell{\;BPA\; \\\cite{bernardini1999ball}} &
\makecell{DeepSDF \\\cite{park2019deepsdf}} &
\makecell{DGP \\\cite{williams2019deep}} & 
\makecell{MIER \\\cite{liu2020meshing}} &
\makecell{PCDLap \\\cite{belkin2009constructing}} &
\makecell{NMLap \\\cite{sharp2020laplacian}} & 
Ours \\
\midrule
\multicolumn{11}{c}{
\includegraphics[width=\columnwidth]{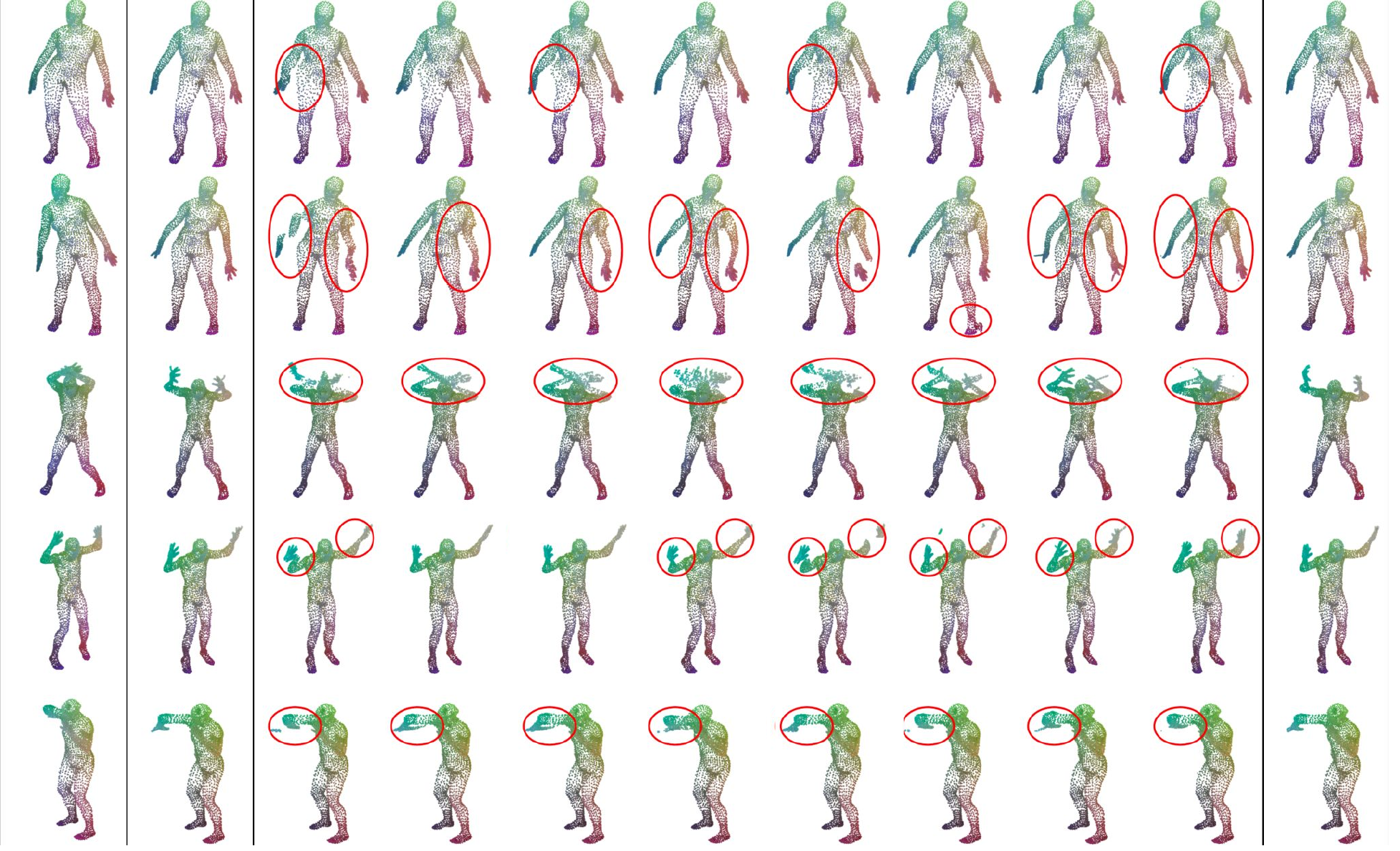}
}
\end{tabular}
}
\vspace{-0.3\baselineskip}
\caption{\textbf{Qualitative comparison of 3D point cloud deformation results on DFAUST~\cite{dfaust:CVPR:2017} dataset (best viewed with 200\% zoom-in).} These deformations are modeled using 32 control handles selected via farthest point sampling.}
\label{fig:dfaust_results}
\vspace{-\baselineskip}
\end{figure}

\begin{figure*}[!t]
\begin{center}
\includegraphics[width=0.8\textwidth, height=0.3\textwidth]{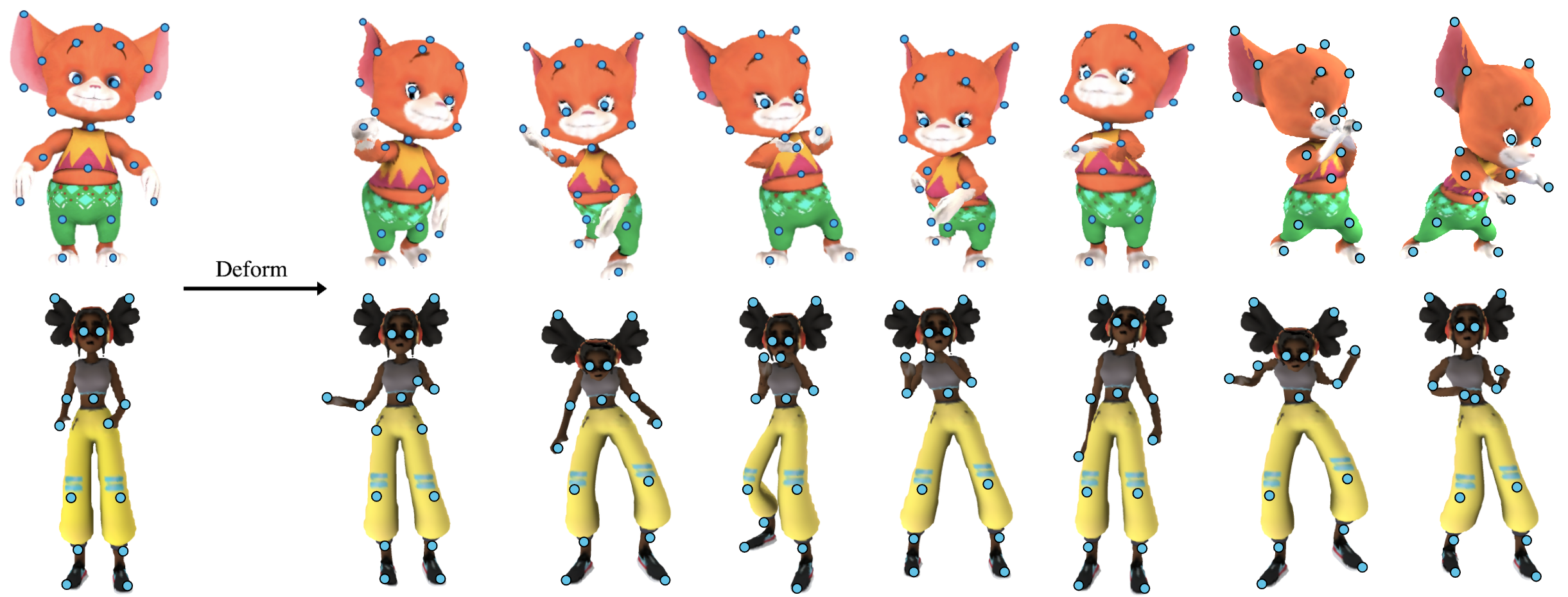} 
\end{center}
\vspace{-1.2\baselineskip}
\caption{\textbf{3D-aware deformation of Mixamo~\cite{mixamo} character images.} 32 control points and their new positions are manually chosen.}
\label{fig:mixamo_results}
\vspace{-0.8\baselineskip}
\end{figure*}

We compare our 3D point cloud deformation results with those computed using the mesh reconstruction methods -- Screened Poisson Surface Reconstruction (PSR) \cite{kazhdan2013screened}, Algebraic Point Set Surfaces (APSS) \cite{apss}, Ball-Pivoting Algorithm (BPA) \cite{bernardini1999ball}, DeepSDF \cite{park2019deepsdf}, Deep Geometric Prior (DGP) \cite{williams2019deep}, and Meshing Point Clouds with Intrinsic-Extrinsic Ratio (MIER). For implicit reconstruction methods that do not directly output a shape in mesh representation, we additionally apply Marching Cubes \cite{lorensen1987marching} to extract a surface mesh. We then convert the surface mesh into a tetrahedron mesh using \cite{hu2020fast} and \cite{huang2018robust} to allow the computation of the shape Laplacian of the underlying \emph{volume}. We also compare our method with point cloud Laplacians -- PCD Laplace (PCDLap) \cite{belkin2009constructing} and Nonmanifold Laplacians (NMLap) \cite{sharp2020laplacian}. In Table~\ref{table:dfaust_results}, our method is shown to outperform all the alternative methods in terms of the three evaluation metrics. This indicates that our method can model handle-based deformations in a more accurate manner by directly \emph{learning} the shape Laplacian from ground truth supervision. In Figure \ref{fig:dfaust_results}, we also provide the visualizations of the deformed point clouds. Our deformations can better preserve the local geometric smoothness of the source shape; they also more accurately match to the ground truth deformations.

\vspace{-0.2cm}
\subsubsection{Experiments on 3D Partial Point Clouds}
We also perform experiments on SHREC'16~\cite{cosmo2016shrec} dataset, which consists of partial shapes of animal objects, such as dog, cat, and wolf. It contains two sub-datasets -- \emph{cuts} and \emph{holes} -- whose names exemplify different types of partiality. As SHREC'16 is originally released as a benchmark for partial correspondence problem, (1) the ground truth \emph{full} shape for each partial shape and (2) their point-wise correspondences are also available. For our experiment, we randomly split the \emph{cuts} and \emph{holes} samples into the training and test sets with an 8:2 ratio.

As our method takes a learning-based approach to estimate the shape Laplacian, it is possible to train our network to predict the shape Laplacian corresponding to the ground truth \emph{full} shape from a \emph{partial} shape. In this way, we can transfer the knowledge of the shape intrinsics of the full shape into the input partial shape to allow more robust shape Laplacian prediction. Similar to the previous experiment, we evaluate our results on the L1 distance between the ground truth and the predicted deformation weights. As shown in Table~\ref{table:partial_point_cloud_results}, our method results in more accurate handle-based deformation weights compared to the case of calculating the shape Laplacian from a mesh constructed by APSS~\cite{apss}, Marching Cubes~\cite{lorensen1987marching} and a volume conversion method~\cite{hu2020fast}.

\begin{table}[!h]
\centering
{ \scriptsize
\setlength{\tabcolsep}{0.2em}
\renewcommand{\arraystretch}{1.0}
\caption{\textbf{Quantitative comparison of 3D partial point cloud deformation results on SHREC'16~\cite{cosmo2016shrec} Dataset}. All weights are computed with respect to 16 point handles selected by farthest point sampling. The ground truth deformation weights are calculated using the topology of the \emph{full} meshes that correspond to partial shapes.}
\label{table:partial_point_cloud_results}

\begin{tabularx}{\columnwidth}{>{\centering}m{4.8cm}|Y}
\toprule
Method & Weight L1 ($\times$100) $\downarrow$ \\
\midrule
APSS~\cite{apss} + Marching Cubes~\cite{lorensen1987marching} & 5.15 \\
Ours & \textbf{4.47} \\
\bottomrule
\end{tabularx}
}
\vspace{-0.5\baselineskip}
\end{table}


\subsection{3D-Aware Image Deformation}
\label{subsec:3d_aware_image_deformation}
We now present our results of 3D-aware image deformation on character and clothed human images.
\vspace{-0.8\baselineskip}

\subsubsection{Experiments on Character Images} 
\vspace{-0.3\baselineskip}
We present our qualitative results of 3D-aware image deformation on \emph{Mousey} and \emph{Michelle} characters in Mixamo~\cite{mixamo} dataset. We remark that images of such characters cannot be manipulated using the existing methods of parametric model fitting \cite{loper2015smpl,zuffi20173d} or human pose transfer \cite{Ma:2017,Li:2019,Chan:2019}. In this experiment, we directly utilize the Laplacian Learning Network trained on DFAUST \cite{dfaust:CVPR:2017} (which is used in Section~\ref{subsec:dfaust_experiments}) and evaluate the model on 3D reconstructions of Mixamo character images -- to demonstrate the \emph{generalizability} of our network across different object categories. However, for 2D-to-3D reconstruction, the pre-trained model of PIFu \cite{saito2019pifu} has not been shown to generalize well to character images. Therefore, we populate separate training images by rendering the Mixamo models and train PIFu from scratch. We create images by rendering the 3D character models with randomly sampled animation frames, camera and light source positions. The total number of images in the dataset is 45,000, and the ratio between the training and test sets is 9:1.

Figure \ref{fig:mixamo_results} shows the results of our 3D-aware image deformation on the images of \emph{Mousey} and \emph{Michelle} characters. Our image deformation method can plausibly model the changes in the pose of the characters \emph{in a 3D-aware manner}. This demonstrates the potential of our method in interactive image editing applications, as it can provide an easy and intuitive interface (i.e., control handles) for image manipulation. We also emphasize again that our Laplacian Learning Network was trained not on Mixamo~\cite{mixamo} but on DFAUST~\cite{dfaust:CVPR:2017} that only consists of human body models. Hence, these results show that our network can be \emph{zero-shot generalizable} to the shapes in other categories by learning the \emph{local} geometric information of a shape. Please note that, while two examples are shown in Figure~\ref{fig:mixamo_results}, more various examples are presented in the supplementary section.

\vspace{-0.2cm}
\subsubsection{Experiments on Clothed Human Images}
\vspace{-0.2\baselineskip}
We also report our results of 3D-aware image deformation on clothed human images. We use RenderPeople~\cite{renderpp} and DeepHuman~\cite{Zheng2019DeepHuman}, which are real-world human model datasets containing textured 3D meshes. Similar to the previous experiment, we prepare the data by rendering the textured 3D human meshes with random viewpoints and lighting. Since RenderPeople provides only 9 models in public, we only show the \emph{qualitative} results using it. We instead use DeepHuman dataset for \emph{quantitative} evaluation, which has approximately 7000 models but with low-quality textures. During the experiment, we again use the Laplacian Learning Network trained on DFAUST~\cite{dfaust:CVPR:2017} dataset, which consists of 3D human body models. For 2D-to-3D reconstruction using PIFu~\cite{saito2019pifu}, we use the official pre-trained model.

\begin{figure}[!h]
\begin{center}
\includegraphics[width=0.95\columnwidth, height=0.5\columnwidth]{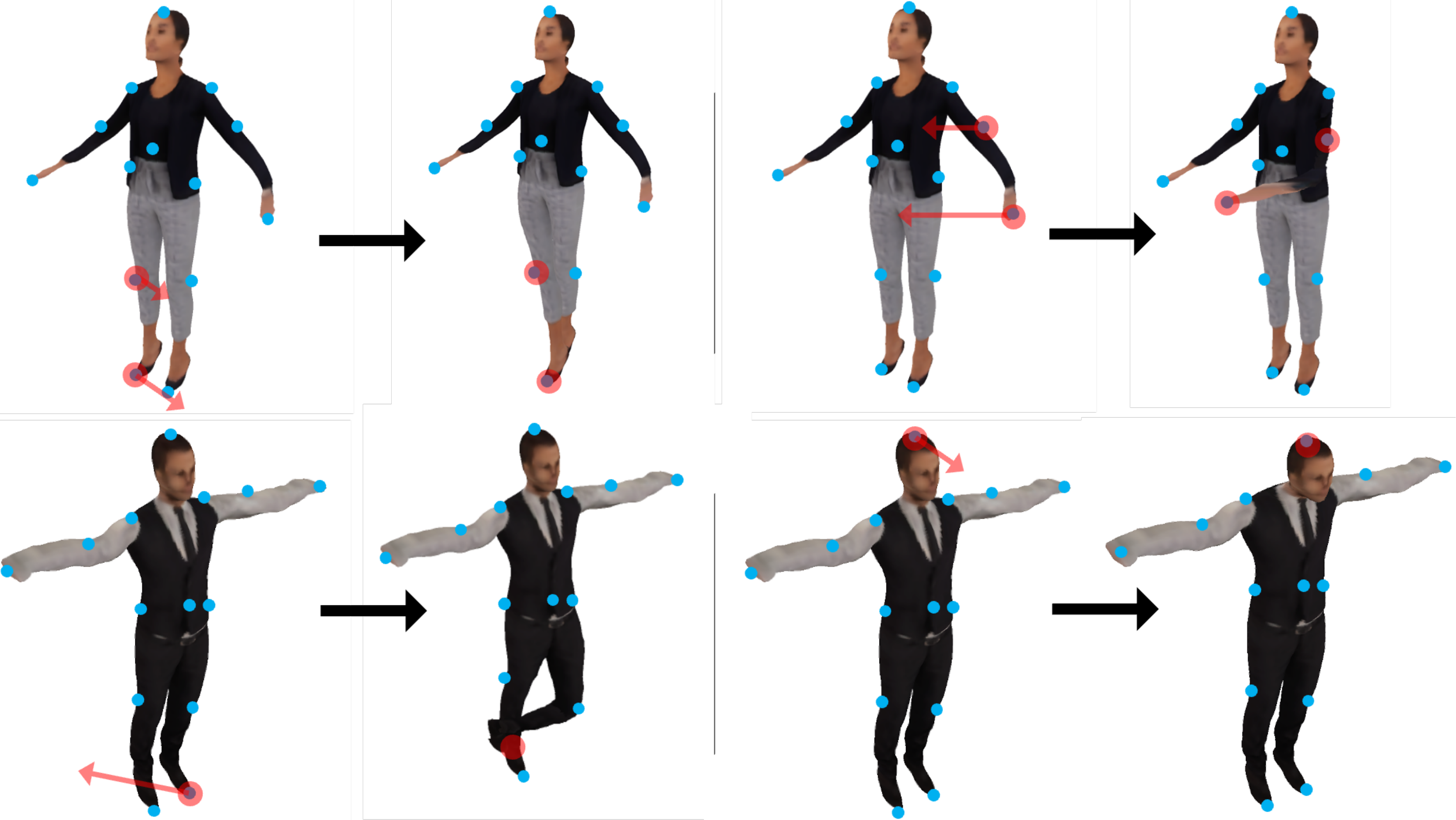} 
\end{center}
\vspace{-0.45cm}
\caption{\textbf{3D-aware deformation of RenderPeople~\cite{renderpp} human images.} Blue dots indicate the user-defined control handles and red arrows show the translations applied to the selected handles.}
\vspace{-0.2cm}
\label{fig:interactive_image_deformation}
\end{figure}

Figures~\ref{fig:teaser_image} and~\ref{fig:interactive_image_deformation} show the qualitative results of 3D-aware image deformation on RenderPeople~\cite{renderpp} dataset. 32 control points are manually picked and manipulated. In Figure~\ref{fig:teaser_image}, our method is shown to produce more plausible image deformations compared to the case of directly computing the shape Laplacian from a mesh reconstructed by PIFu~\cite{saito2019pifu} (whose topology prediction is based on Marching Cubes~\cite{lorensen1987marching}). Figure~\ref{fig:interactive_image_deformation} shows our image deformation examples in user-interaction scenarios. We also provide the quantitative evaluation results of our learned deformation weights on DeepHuman~\cite{Zheng2019DeepHuman} dataset in Table~\ref{tab:image_deformation}. We again compare our results to the deformation weights obtained using the shape Laplcian computed directly from a mesh reconstructed by PIFu. Our method is shown to estimate bounded biharmonic weights in a more accurate manner.

\begin{table}[!h]
\centering
{ \scriptsize
\setlength{\tabcolsep}{0.2em}
\renewcommand{\arraystretch}{1.0}
\caption{\textbf{Quantitative comparison of image deformation results on DeepHuman~\cite{Zheng2019DeepHuman} Dataset}. All weights are computed with respect to the same control handle configurations: 32 point handles selected by farthest point sampling. The ground truth deformation weights are calculated using the geometry intrinsics provided by the ground truth 3D meshes of DeepHuman Dataset.}
\label{tab:image_deformation}

\begin{tabularx}{\columnwidth}{Y|Y}
\toprule
Method & Weight L1 ($\times$100) $\downarrow$ \\
\midrule
Marching Cubes~\cite{lorensen1987marching} & 3.09 \\
Ours & \textbf{2.14} \\
\bottomrule
\end{tabularx}
}
\vspace{-0.5\baselineskip}
\end{table}



\subsection{Ablation Study}
\label{subsec:ablation_studies}
In this section, we perform ablation study to compare different design options for our Laplacian Learning Network. We first explore the effectiveness of the KNN-Based Point Pair Sampling (KPS) and the weight function $\alpha$ that models the sparsity structure of the cotangent Laplacian. As can be seen in Table \ref{table:ablation_study} (columns 2 and 3), we observe significant performance degradation when KPS or $\alpha$ is removed. In the same table (columns 4 and 5), we additionally evaluate the effects of using different symmetric functions for $\gamma_1$ and $\gamma_2$ in the Cotangent Laplacian Prediction Module. We observe that the best performance can be achieved when the proposed functions are used -- in which $\gamma_1$ and $\gamma_2$ are instantiated as absolute difference and element-wise multiplication, respectively. Please refer to the supplementary section for (1) the qualitative results of this ablation study and (2) a comparison with the network that directly learns bounded biharmonic weights -- instead of learning the shape Laplacian.

\vspace{-0.2cm}
\begin{table}[!h]
\centering
{ \scriptsize 
\setlength{\tabcolsep}{0.2em}
\renewcommand{\arraystretch}{1.0}
\caption{\textbf{Ablation study.} $-$KPS and $-\alpha$ indicate settings where KPS or $\alpha$ is removed from our method, respectively. EM Only and AD Only denote settings where $\gamma_1$ and $\gamma_2$ are both instantiated as element-wise multiplication and absolute difference, respectively. All the other experimental setups (e.g., datasets, control handle configurations) are the same as in the DFAUST experiment in Section \ref{subsec:3d_point_cloud_deformation}.}
\label{table:ablation_study}

\begin{tabularx}{\columnwidth}{>{\centering}m{2.8cm}|YYYYY}
\toprule
Metric &
KPS &
{\tiny $-\alpha$} &
EM Only &
AD Only &
Ours \\

\midrule
Weight L1 ($\times$100) $\downarrow$ & 2.79 & 4.95 & 3.54 & 2.35 & \textbf{2.10} \\
Shape CD ($\times$100) $\downarrow$ & 2.46 & 4.07 & 3.27 & 2.20 & \textbf{1.81} \\
Shape HD ($\times$0.1) $\downarrow$ & 0.43 & 1.23 & 0.69 & 0.47 & \textbf{0.42} \\

\bottomrule
\end{tabularx}
}
\vspace{-0.5\baselineskip}
\end{table}


\vspace{-0.2cm}

\section{Conclusion and Future Work}

We propose a method for 3D-aware deformation of 2D images via learning the shape Laplacian. To this end, we introduce a novel network architecture that can learn the shape Laplacian from a 3D point cloud reconstructed from a 2D image. To the best of our knowledge, this is the first study to demonstrate that a learning-based approach can be effective in predicting the shape Laplacian of the underlying volume of a point cloud.

\vspace{-0.4cm}
\paragraph{Negative Societal Impacts}
Our method can be potentially abused to create \emph{deepfake}, while the scope of manipulating images is limited to 3D-aware deformation.  

\vspace{-0.4cm}
\paragraph{Limitations and Future Work}
Since our method models deformations based on bounded biharmonic weights, it cannot guarantee avoiding self-intersections. In addition, the current pipeline learns the shape Laplacian only from the direct supervisions (the L1-losses of the matrices) but not from the outputs of the deformations. We plan to further investigate the ways to improve the deformation quality.

\vspace{-0.4cm}
\paragraph{Acknowledgments.}
We would like to thank Duygu Ceylan for helpful discussions.
This work is in part supported by KAIA grant (22CTAP-C163793-02) funded by the Korea government(MOLIT) and NST grant (CRC 21011) funded by the Korea government(MSIT). M. Sung also acknowledges the support by NRF grant (2021R1F1A1045604) funded by the Korea government(MSIT), Technology Innovation Program (20016615) funded by the Korea government(MOTIE), and grants from the Adobe and KT corporations.

{\small
\bibliographystyle{ieee_fullname}
\bibliography{egbib}
}
\newpage

\newcommand{\manuallabel}[2]{\def\@currentlabel{#2}\label{#1}}
\makeatother
\manuallabel{sec:implementation_details}{3.3}
\manuallabel{subsec:training_details}{4.1}
\manuallabel{table:ablation_study}{4}
\manuallabel{sec:results}{4}
\newcommand{\refpaper}[1]{in the paper}
\renewcommand{\thesection}{S}
\renewcommand{\thetable}{S\arabic{table}}
\renewcommand{\thefigure}{S\arabic{figure}}

\section{Supplementary Material}
\vspace{0.5\baselineskip}

In this supplementary section, we first provide short motion videos generated using our method in Section~\ref{subsec:video}. We then report the additional results of our ablation study in Section~\ref{subsec:additional_ablation_study}. In Section~\ref{subsec:more_implementation_details}, we provide implementation details for the proposed framework. Lastly, we report more qualitative deformation results in Section~\ref{subsec:more_qualitative_results}.

\vspace{0.8\baselineskip}
\subsection{Motion Videos}
\label{subsec:video}

We provide brief descriptions for our video attachment that includes short motion animations created by our image deformation method. Our video, which is available at \url{https://youtu.be/gHxwHxIZiuM}, contains animations of three Mixamo~\cite{mixamo} characters -- \emph{Michelle}, \emph{Mousey}, and \emph{Ortiz} -- in various motions. We also compare our video results to those created using all the alternative methods discussed in Section~\ref{sec:results} of the main paper. 

To create the animations, we deform the rendered images of each Mixamo~\cite{mixamo} character with manually-selected 32 control point handles and their target positions. Note that we use the same experimental setups as in the Mixamo experiment in the main paper (please refer to Section~\ref{sec:results}). In the video, our method is shown to produce 3D-aware image deformations that can express various motions of each character. In addition, our method can generate more plausible deformations in comparison to the alternative methods, which often result in undesired artifacts.

\vspace{0.8\baselineskip}
\subsection{Ablation Study}
\label{subsec:additional_ablation_study}
\vspace{0.3\baselineskip}
\subsubsection{Learning \emph{Deformation Weights}}

We compare our method of learning the shape Laplacian to that of directly learning the deformation weights (i.e., bounded biharmonic weights~\cite{jacobson2011bounded}). To this end, we implement a network that can infer the handle-based deformation weights given a 3D reconstruction represented as a point cloud $\mathcal{P} = \{ \mathbf{p}_i \}_{i = 1 \cdots n}$ and a set of user-defined control points $\{\mathbf{h}_j\}_{j=1,...,m}$\footnote{Among the various types of control handles, we consider \emph{point} handles in this experiment.}. The network consists of three modules: Feature Extraction Module, Control Points Embedding Module, and Weight Regression Module. Feature Extraction Module maps each point $\mathbf{p}_i$ in the input point cloud $\mathcal{P}$ to a feature vector $\mathbf{f}_i$. We use Point Transformer \cite{zhao2021point} architecture as in our original framework. Control Points Embedding Module maps a set of control points $\{\mathbf{h}_j\}_{j=1,...,m}$ to a context vector $\mathbf{c}$ that encodes information about the user handle selection. We use a variant of PointNet \cite{qi2017pointnet} for the module architecture, since Control Points Embedding Module is desired to extract a feature that is permutation-invariant to the input control points. Specifically, we use a shared multilayer perceptron (MLP) network composed of three layers (each of them followed by batch normalization, LeakyReLU, and dropout) and a max pooling layer to aggregate information from all the control points. Next, Weight Regression Module regresses a deformation weight $\mathbf{w}_{j,i}$ associated with the handle $\mathbf{h}_j$ at point $\mathbf{p}_i$, given the following concatenated feature vector:
\vspace{0.4\baselineskip}
\begin{equation}
\mathbf{g}_{ji} = [\mathbf{f}_{\mathbf{h}_j}; \mathbf{c}; \mathbf{f}_i],
\end{equation}

\noindent where $\mathbf{f}_{\mathbf{h}_j}$ denotes a feature vector at the $j$-th point handle (i.e., $\mathbf{h}_j$) that is produced by the Feature Extraction Module. To implement this module, we use the same architecture as that of Cotangent Laplacian Prediction Module, excluding KNN-Based Point Pair Sampling (KPS) and symmetric feature aggregation components. To train our framework, we use L1 losses to inject the direct supervisions for the predicted deformation weights and the intermediate weights, which is analogous to $W$ in our original framework (please refer to Section \textcolor{red}{3.2} in the main paper). In our experimental setting, we specifically train our network to estimate deformation weights for \emph{16 control point handles} that are sampled via \emph{farthest point sampling} to match our test scenario. 

As shown in Table \ref{table:ablation_study_dw}, our original approach of learning the shape Laplacian yields more accurate deformation results compared to that of directly learning the deformation weights. As learning deformation weights is dependent on the control handle selection, we also empirically observed that the deformation weight prediction does not generalize well to a set of control handles whose distribution is different from those of the training examples (e.g., when 16 control points are selected via \emph{random sampling}).

\vspace{0.6\baselineskip}
\begin{table}[!h]
\centering
{ \scriptsize
\setlength{\tabcolsep}{0.2em}
\renewcommand{\arraystretch}{1.0}
\caption{\textbf{Learning the deformation weights vs. learning the shape Laplacian.} All experimental setups are the same as in the DFAUST~\cite{dfaust:CVPR:2017} experiments in the main paper (please refer to Section \textcolor{red}{4.1}). We consider 16 point handles selected via farthest point sampling for both network training and test.}
\label{table:ablation_study_dw}

\begin{tabularx}{\columnwidth}{>{\centering}m{2.8cm}|YYYY}
\toprule
Metric &
Learning $\mathbf{w}$ &
Learning $A$ (Ours) \\

\midrule
Weight L1 ($\times$100) $\downarrow$ & 9.08 & \textbf{2.10} \\
Shape CD ($\times$100) $\downarrow$ & 8.38 & \textbf{1.81} \\
Shape HD ($\times$0.1) $\downarrow$ & 0.44 & \textbf{0.42} \\

\bottomrule
\end{tabularx}
}
\vspace{-0.5\baselineskip}
\end{table}


\vspace{0.1\baselineskip}

\subsubsection{Qualitative Comparisons}

\begin{figure}[!h]
\centering
\setlength{\tabcolsep}{0em}
\def\arraystretch{0.0}
\newcolumntype{Y}{>{\centering\arraybackslash}m{0.125\columnwidth}}
{\tiny
\begin{tabular}{Y|Y|YYYYY|Y}
Input & \makecell{GT\\Deform} 
& \makecell{$-$KPS} &
\makecell{$-\alpha$} &
\makecell{EM Only} &
\makecell{AD Only} &
\makecell{Learning $\mathbf{w}$} &
Ours \\
\midrule
\multicolumn{8}{c}{
\includegraphics[width=\columnwidth, height=0.8\columnwidth]{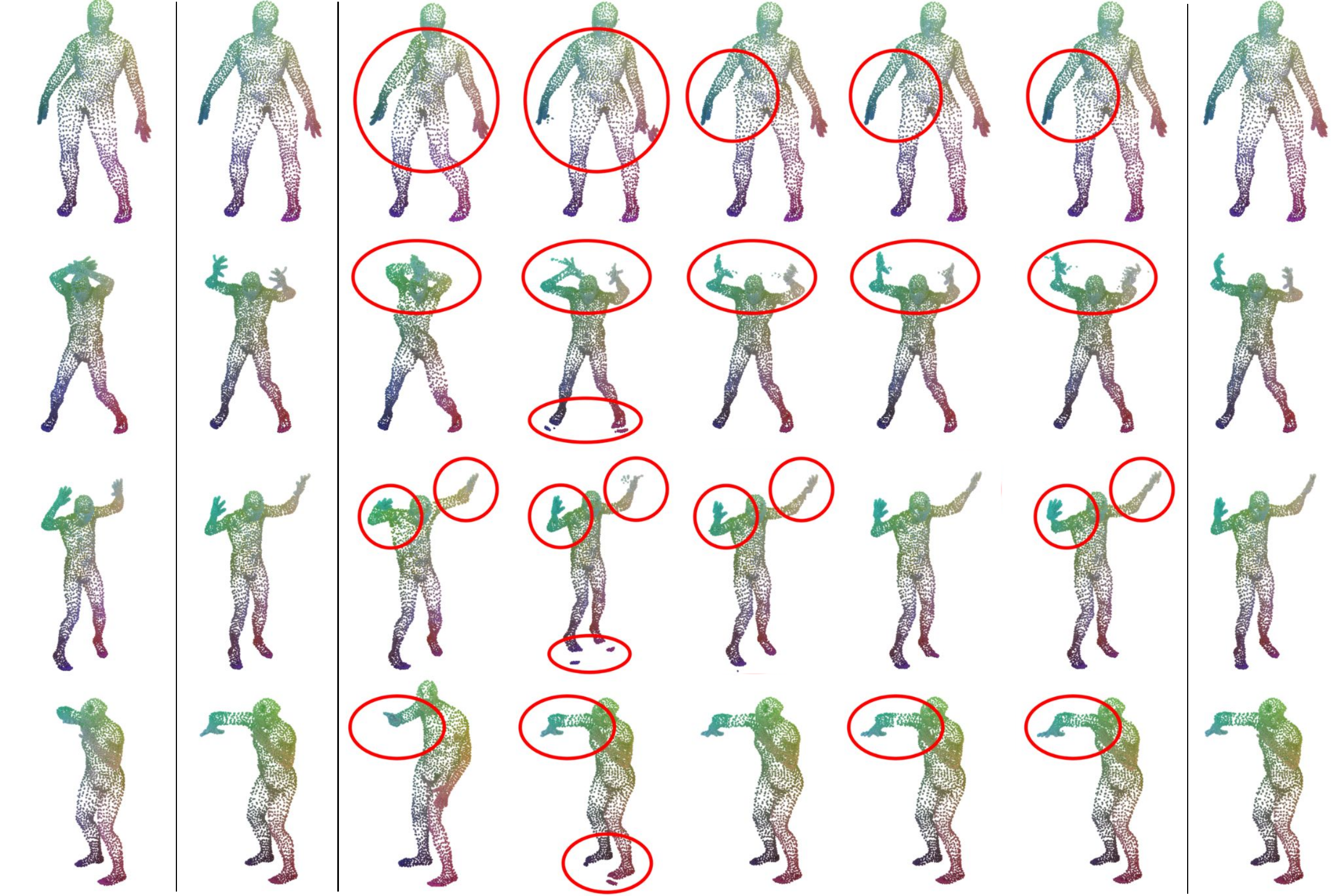}
}
\end{tabular}
}
\caption{\textbf{Qualitative results of our ablation study (best viewed with 200\% zoom-in).} Input denotes the source shape and GT Deform indicates the ground truth deformed shape that is computed using the ground truth shape intrinsics. All experimental setups are the same as in the experiments on DFAUST~\cite{dfaust:CVPR:2017} dataset in Section \textcolor{red}{4.1}. We use 16 FPS point handles for each source shape.}
\label{fig:supp_ablation_qualitative}
\end{figure}

We additionally provide the qualitative results of our ablation study discussed in Tables~\ref{table:ablation_study} and \ref{table:ablation_study_dw}. In Figure \ref{fig:supp_ablation_qualitative}, columns 3, 4, 5 and 6 correspond to the ablation study presented in Table \ref{table:ablation_study} in the main paper. Specifically, $-$KPS and $-\alpha$ indicate settings where KPS or $\alpha$ is removed from our Laplacian Learning Network, respectively. EM Only and AD Only denote settings where $\gamma_1$ and $\gamma_2$ functions in our Cotangent Laplacian Prediction Module are both instantiated as element-wise multiplication and absolute difference, respectively. The column 7 corresponds to the ablation study presented in Table \ref{table:ablation_study_dw} in this supplementary document, where $\textrm{Learning }\mathbf{w}$ denotes the method to directly learn the handle-based deformation weights. Overall, we can observe that the originally proposed method yields the most plausible deformation results than those of the compared settings.



\vspace{0.8\baselineskip}

\subsection{Implementation Details}
\label{subsec:more_implementation_details}
In this section, we report more details of our implementation that could not be included in the main paper due to space limit.
\vspace{-0.1\baselineskip}

\subsubsection{3D Reconstruction Network}
As mentioned in the main paper, we use PIFu \cite{saito2019pifu} to reconstruct an intermediate 3D point cloud from the input image as a prerequisite step to our method. Along with the main PIFu module that learns an implicit function to reconstruct a 3D object geometry, we also train Tex-PIFu module that additionally infers RGB values at given 3D positions of the object surface. Since Tex-PIFu can estimate texture for the object parts that was originally \emph{occluded} in the input image, it allows our method produce image deformations that can \emph{disclose} such occluded parts. When training PIFu and Tex-PIFu, we set the number of epochs as 15 and 10, respectively. We use a batch size of 10 for both modules. Other training setups are the same as those of the original PIFu framework (refer to \cite{saito2019pifu} for more details).

\subsubsection{Network Training} The three modules in our Laplacian Learning Network (i.e., Feature Extraction Module, Cotangent Laplacian Prediction Module, and Inverse Mass Prediction Module) are trained in a joint manner. We use a batch size of 8 and set an initial learning rate to 0.1 with a polynomial learning rate decay schedule. Other training details are the same as in the original Point Transformer segmentation network (refer to \cite{zhao2021point} for more details).

\subsubsection{Metric Computation}

In the experiments on DFAUST \cite{dfaust:CVPR:2017} dataset (please refer to Section \textcolor{red}{4.1} in the paper), we consider three evaluation metrics: (1) L1 distance between the ground truth and the predicted deformation weights, (2) Chamfer distance and (3) Hausdorff distance between the ground truth and the predicted deformed point clouds. Since the ground truth 3D mesh corresponding to each point cloud is available in DFAUST dataset, we first obtain the \emph{ground truth} deformation weights computed using the ground truth topology of the volume mesh generated using \cite{hu2020fast}. In the same manner, we compute the \emph{ground truth} deformed shape using the ground truth deformation weights and the specified control handle configurations. Since our method models a deformation based on a control handle manipulation, we need to specify the initial handle positions and their transformations to generate the deformed shapes for evaluation. To this end, we first sample the initial point handles (with the number of handles specified in Table \textcolor{red}{1}) via farthest point sampling. Given the source shape and the sampled control points, we retrieve the positions of the semantically aligned points in other shapes in the test set by using the shape correspondences provided in DFAUST dataset. We then use them as target control point positions. In our experiment, we randomly generate 4 different deformations for each test shape and use them to evaluate our deformed shape quality.

\onecolumn{
\subsection{Additional Qualitative Results}
\label{subsec:more_qualitative_results}

\subsubsection{Visualization of Learned Deformation Weights}

We visualize our deformation weights learned on the point clouds obtained from DFAUST \cite{dfaust:CVPR:2017} dataset. We show our weights in comparison to those directly computed from the alternative methods (i.e., PSR~\cite{kazhdan2013screened}, APSS~\cite{apss}, BPA~\cite{bernardini1999ball}, DeepSDF~\cite{park2019deepsdf}, DGP~\cite{williams2019deep}, MIER~\cite{liu2020meshing}, PCDLap~\cite{belkin2009constructing}, and NMLap~\cite{sharp2020laplacian}). In Figure \ref{fig:weight_vis}, we show that our method can obtain more accurate deformation weights than the compared scenarios. Especially for the regions around the hands (indicated by red rectangles), our deformation weights are smoothly propagated from the selected control point on the fingertip to \emph{intrinsically} nearby points on the same hand. On the contrary, the compared settings cannot properly distinguish between the two hands. In the same figure, we additionally include the examples of the deformed shapes computed using the shown deformation weights. Our method can produce more plausible shape deformation compared to the other methods.

\begin{figure*}[!h]
\begin{center}
\includegraphics[width=\textwidth]{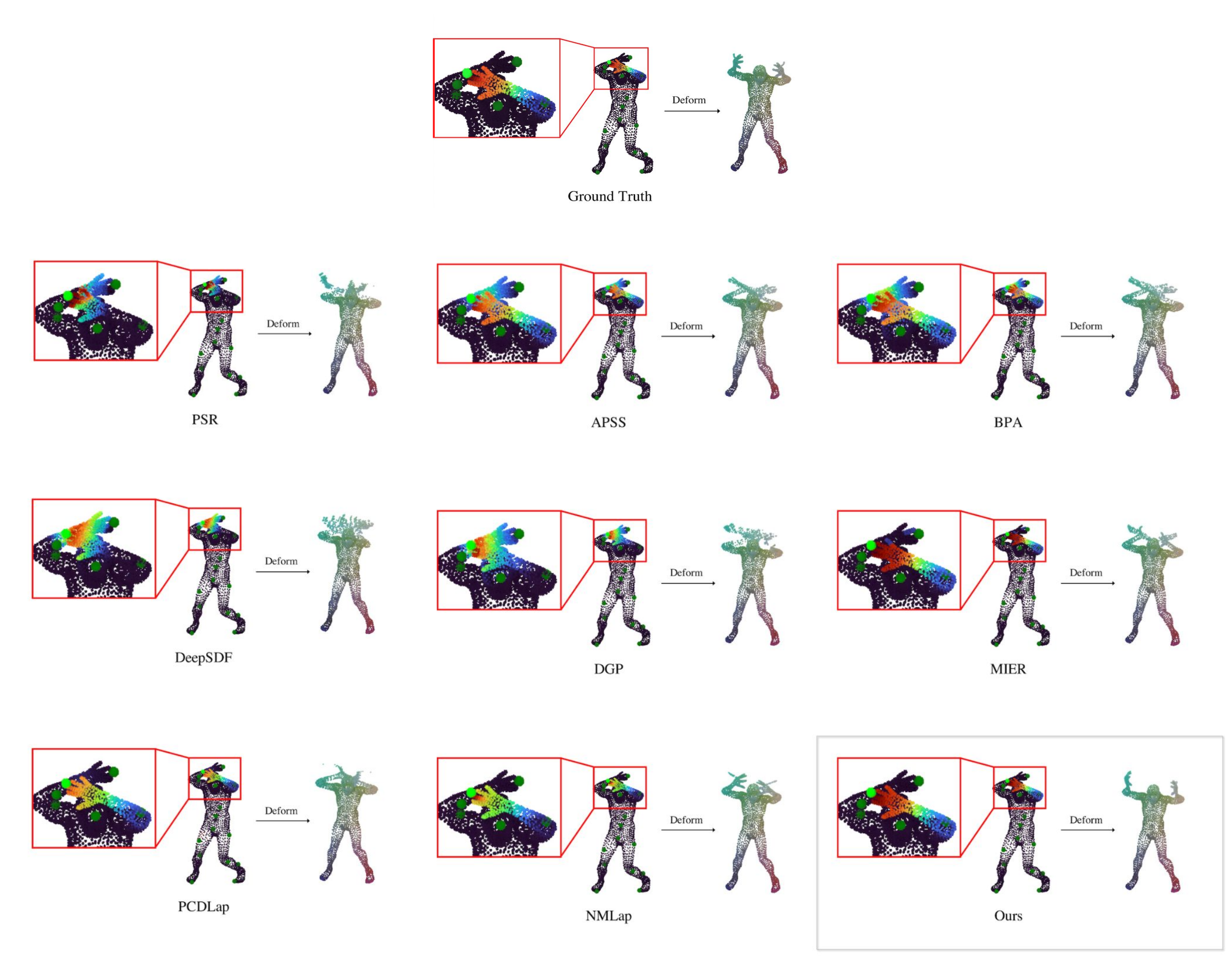} 
\end{center}
\vspace{-\baselineskip}
\captionof{figure}{\textbf{Visualization of Learned Deformation Weights.} Light green spheres indicate the point handles associated with the shown deformation weights, while dark green spheres are the other point handles. The weight intensity is visualized via color coding (i.e., red for high intensity and blue for low intensity). All experimental setups are the same as in the DFAUST~\cite{dfaust:CVPR:2017} experiment in the main paper (please refer to Section \textcolor{red}{4.1}).}
\label{fig:weight_vis}
\vspace{-0.5\baselineskip}
\end{figure*}
}

\onecolumn
\subsubsection{More Comparison Results}
We also provide more qualitative results in comparison to the alternative methods (i.e., PSR~\cite{kazhdan2013screened}, APSS~\cite{apss}, BPA~\cite{bernardini1999ball}, DeepSDF~\cite{park2019deepsdf}, DGP~\cite{williams2019deep}, MIER~\cite{liu2020meshing}, PCDLap~\cite{belkin2009constructing}, and NMLap~\cite{sharp2020laplacian}) discussed in Section \textcolor{red}{4} in the main paper. Specifically, these examples are sampled from our video attachment (please refer to Section~\ref{subsec:video}), which contains the image deformation results of three Mixamo~\cite{mixamo} characters -- \emph{Michelle}, \emph{Mousey}, and \emph{Ortiz}.

\vspace{0.2\baselineskip}
\begin{figure*}[!h]
\begin{center}
\includegraphics[width=0.9\textwidth]{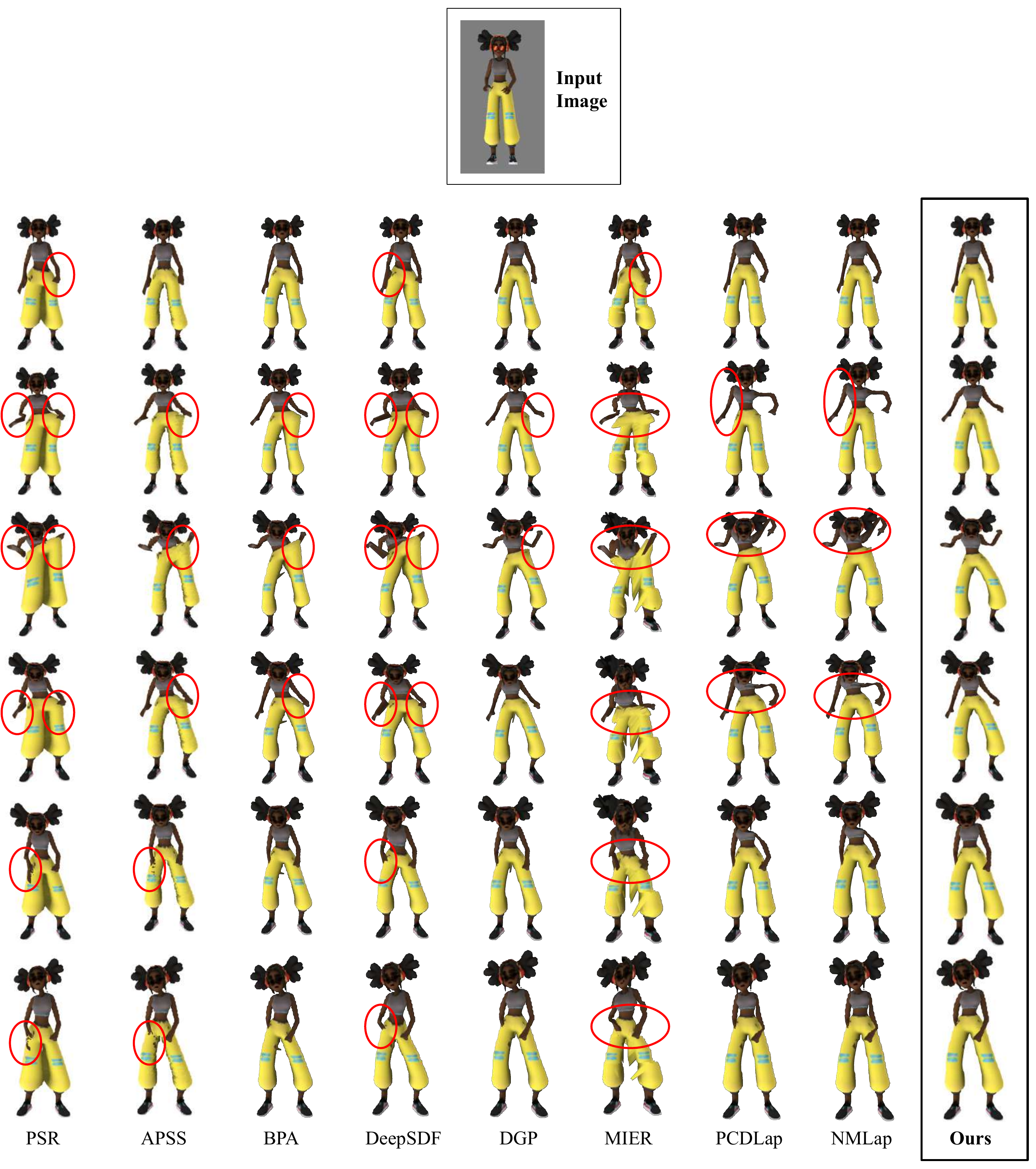}
\vspace{-0.2\baselineskip}
\caption{\textbf{Comparison results on Mixamo~\cite{mixamo} \emph{Michelle} images.} Red circles indicate the image parts with visual artifacts.}
\end{center}
\end{figure*}

\begin{figure*}[!h]
\begin{center}
\includegraphics[width=0.95\textwidth]{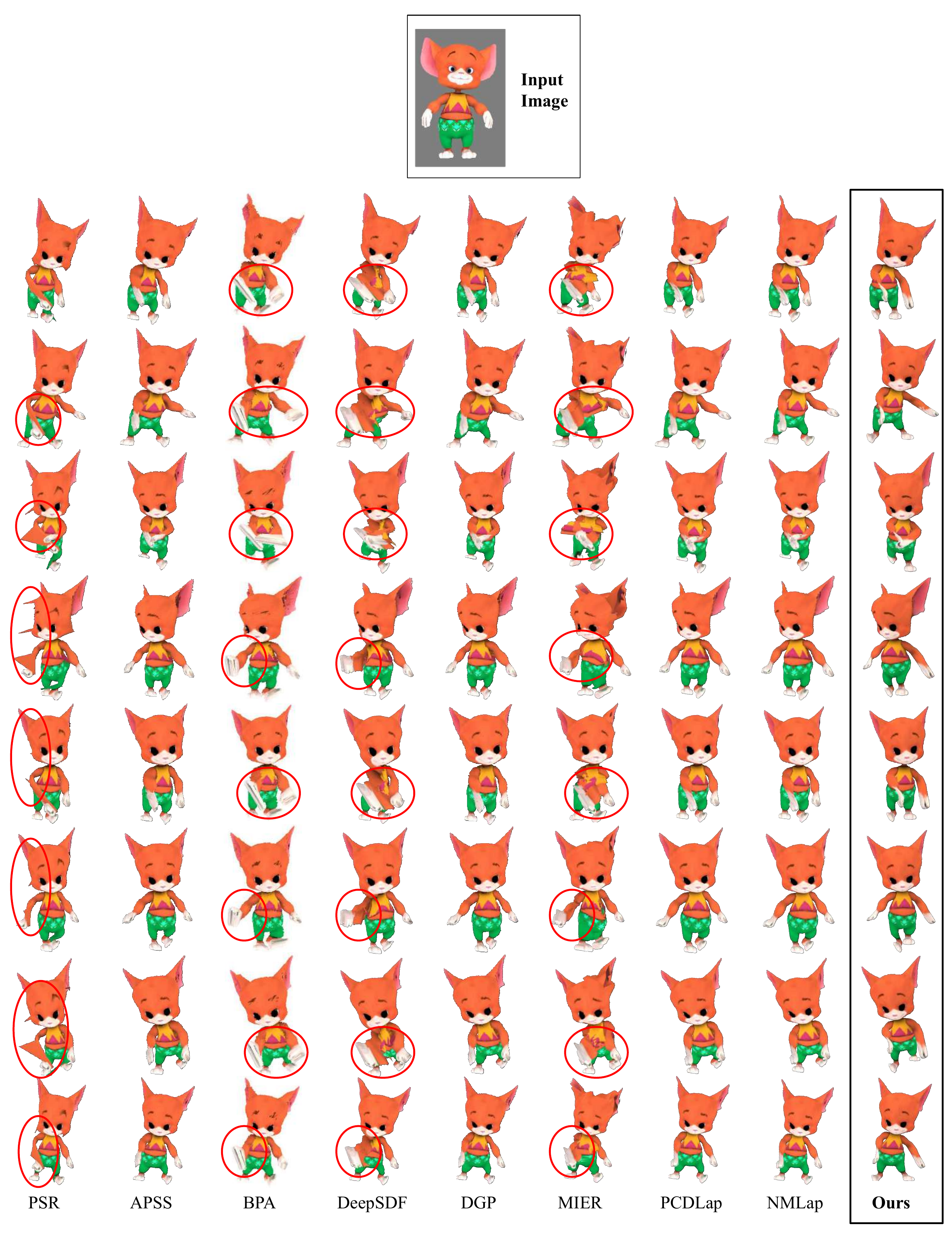}
\vspace{-0.2\baselineskip}
\caption{\textbf{Comparison results on Mixamo~\cite{mixamo} \emph{Mousey} images.} Red circles indicate the image parts with visual artifacts.}
\end{center}
\end{figure*}

\begin{figure*}[!h]
\begin{center}
\includegraphics[width=0.95\textwidth]{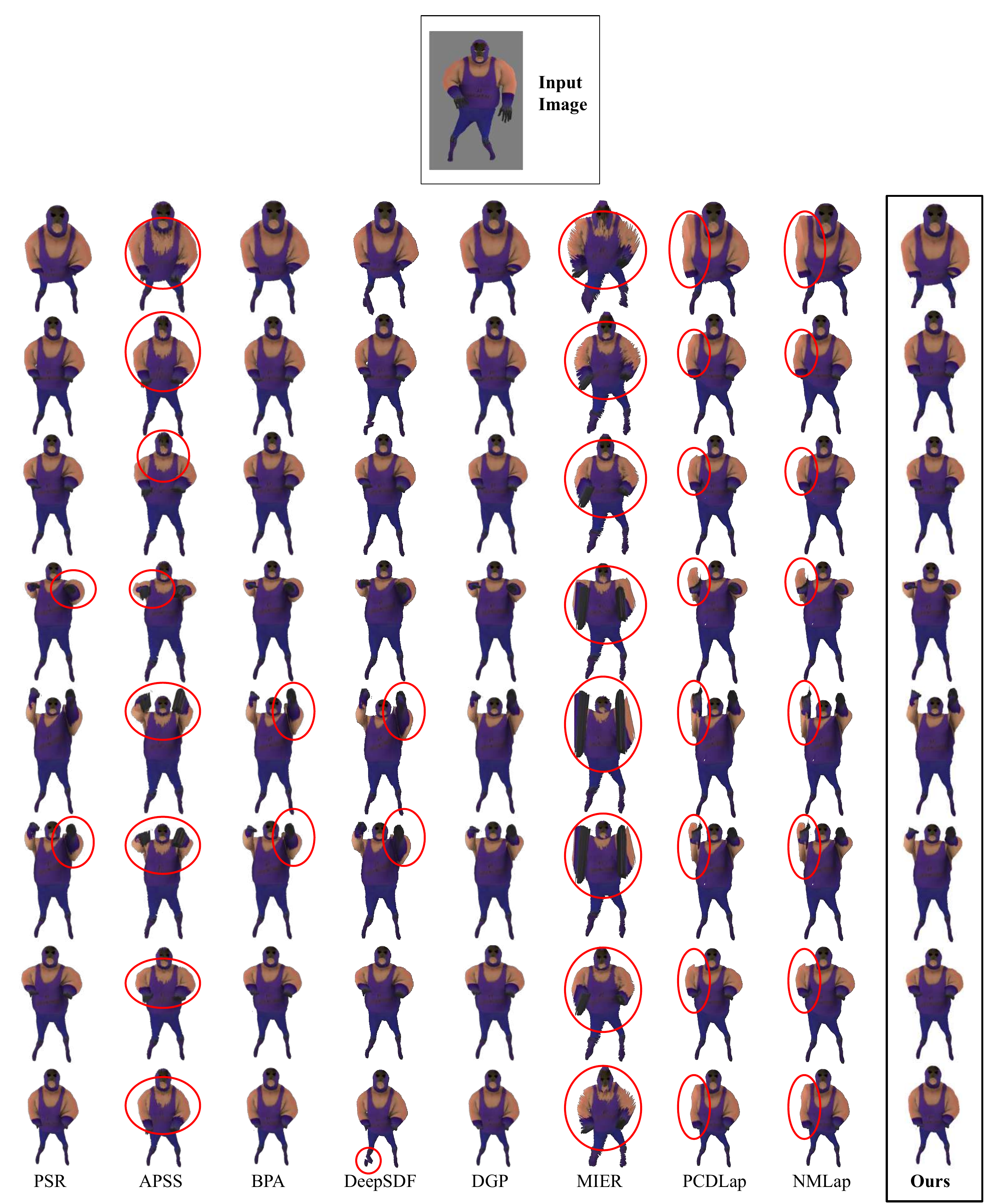}
\vspace{-0.2\baselineskip}
\caption{\textbf{Comparison results on Mixamo~\cite{mixamo} \emph{Ortiz} images.} Red circles indicate the image parts with visual artifacts.}
\end{center}
\end{figure*}


\end{document}